\documentclass[conference]{IEEEtran}
\IEEEoverridecommandlockouts
\usepackage{cite}
\usepackage{amsmath,amssymb,amsfonts}
\usepackage{algorithmic}
\usepackage{graphicx}
\usepackage{textcomp}
\usepackage{xcolor}
\usepackage{float}
\usepackage{algorithm2e}
\usepackage{url}

\def\BibTeX{{\rm B\kern-.05em{\sc i\kern-.025em b}\kern-.08em
    T\kern-.1667em\lower.7ex\hbox{E}\kern-.125emX}}
\begin{document}

\title{Lightweight Decentralized Neural Network-Based Strategies for Multi-Robot Patrolling}

\author{\IEEEauthorblockN{James C. Ward}
\IEEEauthorblockA{
\textit{University of Bristol}\\
Bristol, United Kingdom \\
james.c.ward@bristol.ac.uk}
\and
\IEEEauthorblockN{Ryan McConville}
\IEEEauthorblockA{
\textit{University of Bristol}\\
Bristol, United Kingdom \\
ryan.mcconville@bristol.ac.uk}
\and
\IEEEauthorblockN{Edmund R. Hunt}
\IEEEauthorblockA{
\textit{University of Bristol}\\
Bristol, United Kingdom \\
edmund.hunt@bristol.ac.uk}
}

\maketitle

\vspace{-5mm}

\begin{abstract}
The problem of decentralized multi-robot patrol has previously been approached primarily with hand-designed strategies for minimization of ``idlenes'' over the vertices of a graph-structured environment. Here we present two lightweight neural network-based strategies to tackle this problem, and show that they significantly outperform existing strategies in both idleness minimization and against an intelligent intruder model, as well as presenting an examination of robustness to communication failure. Our results also indicate important considerations for future strategy design.
\end{abstract}

\section{Introduction}

\subsection{The multi-robot patrolling problem}

Robotic surveillance for infrastructure security or environmental monitoring is of considerable interest in the multi-robot systems community~\cite{Bayat2017, Raty2010}. The most common approach for modelling patrol problems with ground-based agents is to represent points of interest to be observed as the vertices $\mathcal{V}$ of a weighted undirected patrol graph $\mathcal{G}(\mathcal{V},\mathcal{E})$~\cite{Machado2022}. The ``idleness''~\cite{Machado2022} or ``latency'' of $\mathcal{V}$ is the most common criterion used to measure performance, where both terms are defined as the time since the last visit of a patrol agent to a vertex. The problem of multi-agent idleness minimization over $\mathcal{G}$ has seen significant attention in the literature~\cite{Basilico2022}, with a wide range of both centralized and decentralized strategies presented. In this work, we present two novel decentralized strategies for idleness minimization in the multi-robot patrol problem. The first strategy utilizes a lightweight Graph Neural Network (GNN)-based priority function to operate on the patrol graph, and inter-agent coordination techniques inspired by existing strategies. The second strategy utilizes an extremely minimal priority function based on a regression of the GNN-based strategy, and the same inter-agent coordination. We show that both of our new strategies significantly outperform leading existing strategies in both idleness minimization and adversarial patrolling.

\subsubsection{Centralized versus decentralized patrol}

The problem of minimizing average or maximum idleness over a patrol graph $\mathcal{G}$ with $k$ agents is fundamentally a variant of the Traveling Salesman Problem (TSP), and is known to be NP-hard in the single-~\cite{Papadimitriou1977} and multi-agent~\cite{Pasqualetti2011} cases. For $k=1$, the optimal strategy for idleness minimization is simply to follow the solution to the TSP on $\mathcal{G}$~\cite{Chevaleyre2004}. For $k>1$, \cite{Chevaleyre2004} separates possible solutions into two categories -- ``cyclic'' strategies, in which all agents are spaced along and follow the same path, found by solving the TSP on $\mathcal{G}$, and ``partition-based'' strategies, in which $G$ is partitioned into $k$ disjoint subgraphs, and each agent following the TSP solution of its respective subgraph. Work by \cite{Afshani2020}\cite{Afshani2022} extends this to the case where $G$ can be partitioned into $l\leq k$ subgraphs, with subgraphs possibly containing multiple agents, and presents a polynomial-time approximation algorithm for the optimal solution to minimizing maximum idleness. \cite{Pasqualetti2011} also present an approximation algorithm, alongside an analysis of the computational complexity of the problem. The case where \textit{weighted}\ idleness, i.e. node idleness multiplied by a node weight, is the desired minimization criterion has also been examined for single-~\cite{Alamdari2014} and multi-agent~\cite{Afshani2020} cases.

While these approaches can offer solutions within a proven approximation factor of optimality~\cite{Afshani2022}, they assume that $\mathcal{G}$ is both fixed and known, and are intolerant to both agent attrition and the addition of new agents. A centralized monitoring and control system would be required to react to any change in the environment or patrol system, and as the time complexity of any centralized strategy scales at least polynomially with the degree $n$ of $\mathcal{G}$\footnote{Lower bound limited by the time complexity of the TSP solver used -- popular algorithms include Christofides' algorithm ($\mathcal{O}(n^3)$)~\cite{Christofides1976}, Held-Karp ($\mathcal{O}(2^nn^2)$)~\cite{HeldKarp1962}, 2-opt ($\mathcal{O}(n^2)$)~\cite{Croes1958}, and Lin-Kernighan ($\geq\mathcal{O}(n^2$)~\cite{Lin1973}. Most commonly used high-performing TSP solvers have complexity of at least $\mathcal{O}(n^2)$.} (for partition-based solutions, significant additional scaling is present\footnote{The highest performing approximate algorithm~\cite{Afshani2022} scales with the number of agents $k$ as $\mathcal{O}((k/\epsilon)^k)$, significantly reducing the practicality of solving for large team sizes.}), they could become impractical to calculate for large environments or teams. As such, decentralized strategies are highly desirable for real-world deployments due to increased practicality and robustness, significantly reduced time to deployment, and no need for established communication networks. It has also been found that decentralized communication can improve accuracy of consensus in uncertain environments in patrol scenarios~\cite{madin:communication}. Consequently, decentralized patrol has seen significant attention over recent years~\cite{Huang2019}\cite{Basilico2022}, with numerous strategies presented in the literature~\cite{Machado2022, Yan2016, Almeida2004}. Notably high-performing strategies include the Bayesian family of strategies presented by \cite{Portugal2013_2, Portugal2013, Portugal2016} and the dynamic task assignment family presented by \cite{Farinelli2017}.

\subsubsection{Adversarial patrol}

Multi-robot patrol is also often considered in the context of the patrol team trying to prevent an intelligent adversary from gaining undetected access to the environment. In the case of a single adversary and single patrol agent, this can be tackled either with game theoretical approaches, in which the problem can be modeled as a Bayesian Stackelberg game~\cite{Paruchuri2004}\cite{Yang2019}, or by defining the movement of the agent as a Markov chain on the patrol graph and optimizing for the probability of detecting an adversary~\cite{Yang2019}\cite{Duan2021}. Fence patrol, in which the patrol graph is a single line or closed loop, has also seen considerable examination in adversarial contexts, as the heavily constrained environment allows direct optimization of performance in cases such as multiple adversaries carrying out coordinated simultaneous attacks~\cite{Sless2014} or sequential attacks~\cite{Sless2019}. 

However, the adversarial performance of multi-agent teams on an arbitrary patrol graph is a more complex problem and cannot easily be tackled directly. Predictable patrol strategies are at an obvious disadvantage when facing an intelligent adversary~\cite{Agmon2011}, and to our knowledge no decentralized multi-agent strategies have been presented that are directly designed for adversarial performance. \cite{Ward2023} present an empirical method to measure the performance of simulated patrol strategies against various adversary models and demonstrate that idleness-based performance measures do not always adequately predict adversarial performance.

\subsection{Graph neural networks}
Graph neural networks (GNNs)~\cite{Zhou2020} are a method of applying deep learning principles to graph-structured data. The fundamental principle of GNNs is the ``graph convolution'', whereby information propagates between connected vertices of a graph, allowing for a similar process to standard convolution on unstructured data. Iterated graph convolution then allows for vertex data to propagate throughout a graph, as $k$ 1-hop convolutions allow for the effective ``perceptual range'' of each vertex to extend up to $k$-hops away. The GNN used in this work, discussed in detail in Section \ref{section:SUN}, uses a graph convolutional network (GCN) based architecture modified to account for edge information, inspired by the method used by \cite{Hart2020} (which is itself based on a method presented by \cite{Battaglia2018}).

\subsection{Learning-based approaches to patrolling}

GNNs present an obvious model for learning behavior in graph-structured environments and can also be applied to communication graphs within teams of robots, allowing for learning multi-robot coordination mechanisms~\cite{tolstaya:gnn}. As such, the potential of GNN-based methods for variations of the multi-robot patrolling problem has seen some investigation. \cite{tolstaya:gnn_patrol} apply this to the multi-robot coverage problem, in which a team of robots must efficiently visit every vertex of interest in a graph within a time limit. \cite{guo:marl_approach} present a centralized GNN-based controller which attempts to optimize for unpredictability for adversarial patrol scenarios, however the controllers learned by this method are environment-specific and do not generalize to varying team sizes. In this work, we do not leverage GNNs for inter-robot coordination, instead only using them to act on the graph-structured patrol environment. The reasons for this are discussed in Section \ref{section:discussion}.

\section{Problem definition}

For a given environment that we wish to monitor, we select a set of points from which it can be observed by an agent. These points can be selected either by choosing specific regions that require monitoring, or simply by partitioning the environment such that the entire environment can be observed from these points, based on the perceptual range of the agent. These points become the vertices $\mathcal{V}$ of a weighted undirected patrol graph $\mathcal{G}$, where the edges $\mathcal{E}$ of $\mathcal{G}$ are the traversable routes between members of $\mathcal{V}$, with weights corresponding to their distances. For the efficient monitoring of the environment, our goal is to use a team of mobile agents to minimize the average instantaneous idleness of $\mathcal{V}$ across an arbitrary time period, where ``idleness'' is defined as the time since a vertex was last visited by an agent. To achieve this, we propose a decentralized control strategy for the agents, whereby each agent maintains an internal belief of $\mathcal{G}$ and the idlenesses of $\mathcal{V}$, and makes its decisions based on its own observations and any messages received from other agents. In principle, inter-agent communication could be range-limited, require line-of-sight, or any other limitation, and as such we present an examination of the robustness of our proposed strategies to imperfect communication later in this work. However, for the main part of this work we assume unobstructed communication, thus allowing every agent's belief of $\mathcal{G}$ to be equal to its true state.



For the problem of adversarial patrol, we consider the case of an attacker that, having made the decision to attack a given node at a given time, is successful if no patrol agent visits that node for a specified amount of time. In our tests we implement this using the ``intelligent adversary'' model proposed by \cite{Ward2023}, which attempts to optimize its probability of successfully attacking based on observations of the patrol agents and environment.

\section{The Spatial Utility Network Strategy}
\label{section:SUNS}

We propose a decentralized, neural network-based controller for the multi-robot patrolling problem, called the Spatial Utility Network Strategy (SUNS)\footnote[3]{Our implementations of both SUNS and MNS are available at \url{https://github.com/jward0/patrolling_sim}}. As it does not attempt to generate centralized policies or graph-level embeddings on the patrol graph, it can function with dynamic graphs, variable team sizes, and constrained communications. All of the testing discussed later in this work was carried out with the same trained controller instance on every agent with no modification for team size or environment, demonstrating this flexibility. 

SUNS for each agent is described in full in Algorithm \ref{alg:suns}, where $\mathcal{V}$ refers to the vertices of the patrol graph, and $\mathcal{N}_{current}$ refers to the neighbors of the vertex the agent is at. The utility function $U$ is described in full in Section \ref{section:SUN}. \footnote[4]{It should be noted that this controller only serves to determine high-level navigation goals for a robot, and any behaviors relating to collision avoidance, local path planning, or calling to recalculate targets after navigation failure should be handled by an appropriate navigation stack --- our simulations (see Section~\ref{sec:testing}) used the ROS navigation stack (http://wiki.ros.org/navigation) with AMCL (http://wiki.ros.org/amcl)}

\RestyleAlgo{ruled}
\SetKwComment{Comment}{/* }{ */}
\begin{algorithm}
\caption{SUNS description}
\label{alg:suns}
idleness = [0 for $v$ in $\mathcal{V}$]\;
intentions = empty\;
$v_n$ = start node\;
\While {true} {
    utilities = [0 for $v$ in $\mathcal{V}$]\;
    \For{$i \in \mathcal{N}_{current}$} {
        utilities$[i] = U(i)$\;
    }
    \eIf {all $i \in \mathcal{N}_{current}$ is in intentions} {
        \texttt{nothing}\;
    } {
        \For {$i \in$ intentions} {
            utilities$[i] = 0$\;
        }
    }
    $v_{n+1}$ = \texttt{argmax}(utilities)\;
    \texttt{broadcast}[$v_n$, $v_{n+1}$]\;
    \texttt{start\_move to} $v_{n+1}$\;
    \While {moving to $v_{n+1}$} {
        \texttt{update\_idleness} 
        \If{broadcast received} {
            \texttt{update\_idleness} 
            \texttt{update\_intentions} 
        }
    }
    $v_n$ = $v_{n+1}$\;
}

\end{algorithm}

\subsection{Spatial Utility Network}
\label{section:SUN}

The Spatial Utility Network (SUN) is the utility function used by SUNS, consisting of a GNN acting on a patrol graph, pre-trained using advantage actor-critic (A2C)~\cite{Mnih2016} reinforcement learning with separate instances of the SUN acting as both actor and critic.

\subsubsection{Architecture}
For a vertex $i$ with associated component $\textbf{v}_i$ of an arbitrary graph signal connected to neighbors $\mathcal{N}_i$ by edges with arbitrary edge values (i.e. a vector of values associated with an edge) $\textbf{e}_{ij}$, the utility $u_i$ is calculated as:

\begin{equation}
    u_i = U(i) = f_1(\textbf{v}_i) + \sum_{j\in \mathcal{N}_i}f_2([\textbf{v}_j | \textbf{e}_{ij}])
\end{equation}

Where $[\textbf{v}_j | \textbf{e}_{ij}]$ represents the concatenation of  $\textbf{v}_j$ and $\textbf{e}_{ij}$ into a single vector $[\textbf{v}_{j, 0}, \textbf{v}_{j, 1}, ... \textbf{e}_{ij, 0}, \textbf{e}_{ij, 1}...]$, and $f_1$ and $f_2$ are multi-layer perceptrons operating on self- and neighbor-information respectively. This structure is equivalent to graph convolution (with the unweighted adjacency matrix $\textbf{A}$ as the graph shift operator) layered on top of the two perceptrons. This structure allows for explicit learning of a response to not only the structure of the graph, but also information encoded in its edges. Centralizing this calculation (i.e. applying simultaneously to the entire set $\mathcal{V}$ of $N$ vertices) gives:

\begin{equation}
    U(\mathcal{V}) = f_1\left(\begin{bmatrix} \textbf{v}_0 \\ \vdots \\ \textbf{v}_N \end{bmatrix}\right) + \textbf{A}f_2\left(\begin{bmatrix} [\textbf{v}_0 \textbf{e}_{00}] & \hdots & [\textbf{v}_0 \textbf{e}_{N0}]\\ \vdots & & \vdots \\ [\textbf{v}_N \textbf{e}_{0N}] & \hdots & [\textbf{v}_N \textbf{e}_{NN}]\end{bmatrix}\right)
\end{equation}

Where $f_1$ and $f_2$ are broadcast to each element of their input tensors. This calculation is performed independently on each robot in a multi-agent scenario.

In this work, each graph signal component $\textbf{v}_i \in \mathbb{R}^2$ is a pair of values consisting of the instantaneous idleness of vertex $i$ and the distance of the shortest path between the agent and vertex $i$. As the SUN is only called when an agent is at a vertex, these shortest paths can be calculated ahead of time and only recalculated if the graph changes. The edge values $\textbf{e}_{ij}$ consist of only the weight of the edge connecting vertices $i$ and $j$.

Due to the small input space of $f_1$ and $f_2$ for this problem, the networks themselves are similarly lightweight. Both are 3-layer fully connected perceptrons, with leakyReLU $(a=0.3)$ as the activation function. $f_1$ has an input layer of size 2, a hidden layer of size 4, and an output layer of size 1. $f_2$ has an input layer of size 3, a hidden layer of size 6, and an output layer of size 1.

Due to the GNN structure of the SUN, its perceptual range and depth can be increased by stacking multiple instances of a SUN layer on top of each other. If we treat the observed idlenesses as the initial utilities $u_0$, then updated utilities $u_k$ with perceptual range $k$ are calculated by passing the vector of vertex values through the SUN $k$ times, each time updating the utility values to the most recently calculated values. In this work, we found that $k=1$ gave the best performance against our validation dataset during training, and we therefore used this for our final model.

\subsubsection{Training}

The SUN was trained on a single agent using A2C reinforcement learning. The actor comprised a SUN instance feeding into a softmax layer, and the critic comprised a separate SUN instance feeding into a 1D max-pooling layer. The training environment was a Euclidian patrol graph with the agent constrained to move along its edges. Upon arrival at a vertex, the agent received reward proportional to the instantaneous idleness of the vertex. The corresponding action space is the set of vertices of the patrol graph, where selecting a vertex $v$ as an action will cause the agent to step once along the shortest path towards $v$. The state space is the inputs to the SUN, i.e. the weighted adjacency matrix of the patrol graph and the associated ``vertex values''. The agent was trained sequentially on 10 randomly generated graphs with a range of numbers of vertices (15-80), number of edges (15-130), minimum edge lengths (2-12m), and maximum edge lengths (3-25m). Our validation set comprised 8 more randomly generated graphs and two real patrol graphs taken from \textit{ROS patrolling sim} (see Section \ref{sec:patrolsim}) that would not be used in final testing. 30 SUN instances were initialized with random weights and trained, and the instance which performed the best on our validation set was selected and incorporated into the full SUNS architecture for testing. This trained instance is present in our provided implementation.

\subsection{Multi-robot coordination}

\begin{figure}[]
    \centering
    \normalsize
     \includegraphics[width=0.93\linewidth]{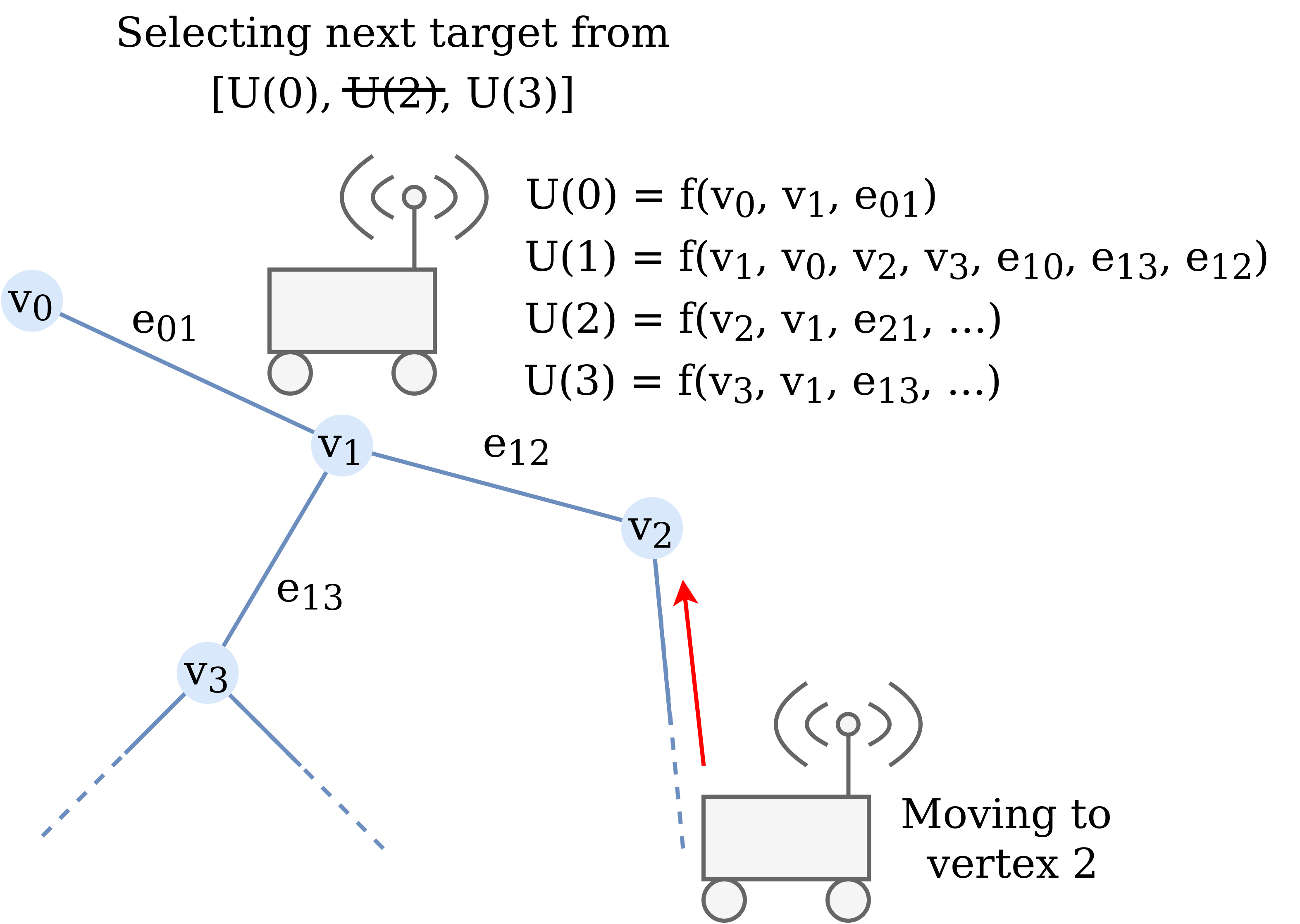}
    \centering
    \vspace{-2mm}
    \caption{Overview of multi-robot patrol decision making and coordination.}
    \vspace{-6mm}
    \label{fig:patrol_overview}
\end{figure}

Extending the trained SUN to a multi-robot controller required the addition of coordination rules. Similarly to the SEBS algorithm~\cite{Portugal2013}, when an agent arrives at its target vertex, it calculates the utility of all adjacent vertices. Any vertex that another agent has announced it is traveling to is disregarded, and the remaining vertex with the largest utility is selected as its next target. If all adjacent vertices have been announced as targets by other agents, the agent skips this process and selects the vertex with the highest utility as its next target. Once the agent has selected its next target, it broadcasts its current and target vertices to other agents, so they can update their idleness logs and account for agents' intentions in their decision making process. This process is illustrated in Figure~\ref{fig:patrol_overview}.

\subsection{The minimal network}
During testing of the SUN, it was found that removing the GNN elements -- i.e. only leaving the perceptron acting on each node in isolation -- had only a small impact on performance. Further testing revealed that the remaining perceptron could be reduced to only three neurons with again only a minor change in performance. The remaining strategy -- a three-neuron utility function and a single simple rule for multi-robot coordination -- represents an extremely minimal strategy for decentralized patrol, while offering a high level of performance. We refer to this strategy as the Minimal Network Strategy (MNS)\footnote[3]{Our implementations of both SUNS and MNS are available at \url{https://github.com/jward0/patrolling_sim}}  throughout the rest of this work. 

\section{Testing}
\label{sec:testing}
We assess the performance of the new SUNS and MNS strategies in simulation, in both idleness minimization and defense against intelligent attackers, in comparison to three other leading strategies.
\vspace{-1mm}
\subsection{ROS patrolling sim}
\label{sec:patrolsim}

\textit{ROS patrolling sim}\footnote[5]{\url{http://wiki.ros.org/patrolling_sim}}~\cite{Portugal2018} is a simulator created for the testing and benchmarking of multi-robot patrol strategies. While some sim-to-real gap is inevitable, it includes lidar, odometry, path-planning, and navigation in a fully simulated 2.5D environment, allowing for both a high level of confidence that it can meaningfully simulate a real environment and portability of any patrol strategies developed inside the simulator to real robots. The final testing for this work was carried out in \textit{ROS patrolling sim}, allowing it to be well-placed within the literature -- the strategies proposed by \cite{Yan2016}, \cite{Portugal2013}\cite{Portugal2016}, and \cite{Farinelli2017} were all tested in this simulator and successfully ported to real robots. It is worth noting that while this presents an idealized environment, there is still sufficient noise in the simulation that repeated trials from identical initial conditions will result in different trajectories being followed by the robots once enough time has passed, and consequently different steady-state behavior and performance.
\vspace{-1mm}

\subsection{MAGESim}
\vspace{-1mm}
\label{sec:magesim}
\textit{Multi-Agent Graph Environment Simulator (MAGESim)}\footnote[6]{\url{https://github.com/jward0/magesim}} is a simulator originally created as a light-weight, high-speed alternative to \textit{ROS patrolling sim}, designed to be easily customizable to any specific use-case regarding agents acting in a graph-structured environment. As it presents a highly constrained environment with no simulation of lidar, odomoetry, collisions, or realistic path-planning \textit{ROS patrolling sim} remains generally preferable for patrolling performance analysis. However, as \textit{MAGESim} offers much higher simulation speed\footnote[7]{Up to 2500 times faster than \textit{ROS patrolling sim} on the author's hardware, depending on patrol strategy}, it is useful for rapid development and assessment of novel strategies, or for performing performance sweeps across large ranges of parameters. As such, in this work our examination of robustness to imperfect communication was carried out in \textit{MAGESim}, as varying communication failure rate added an extra dimension to our parameter space beyond what was considered for our main performance analysis.

\vspace{-1mm}

\subsection{Test protocols}
\vspace{-1mm}
\subsubsection{Main performance analysis}
\begin{figure*}
    \includegraphics[width=0.75\linewidth]{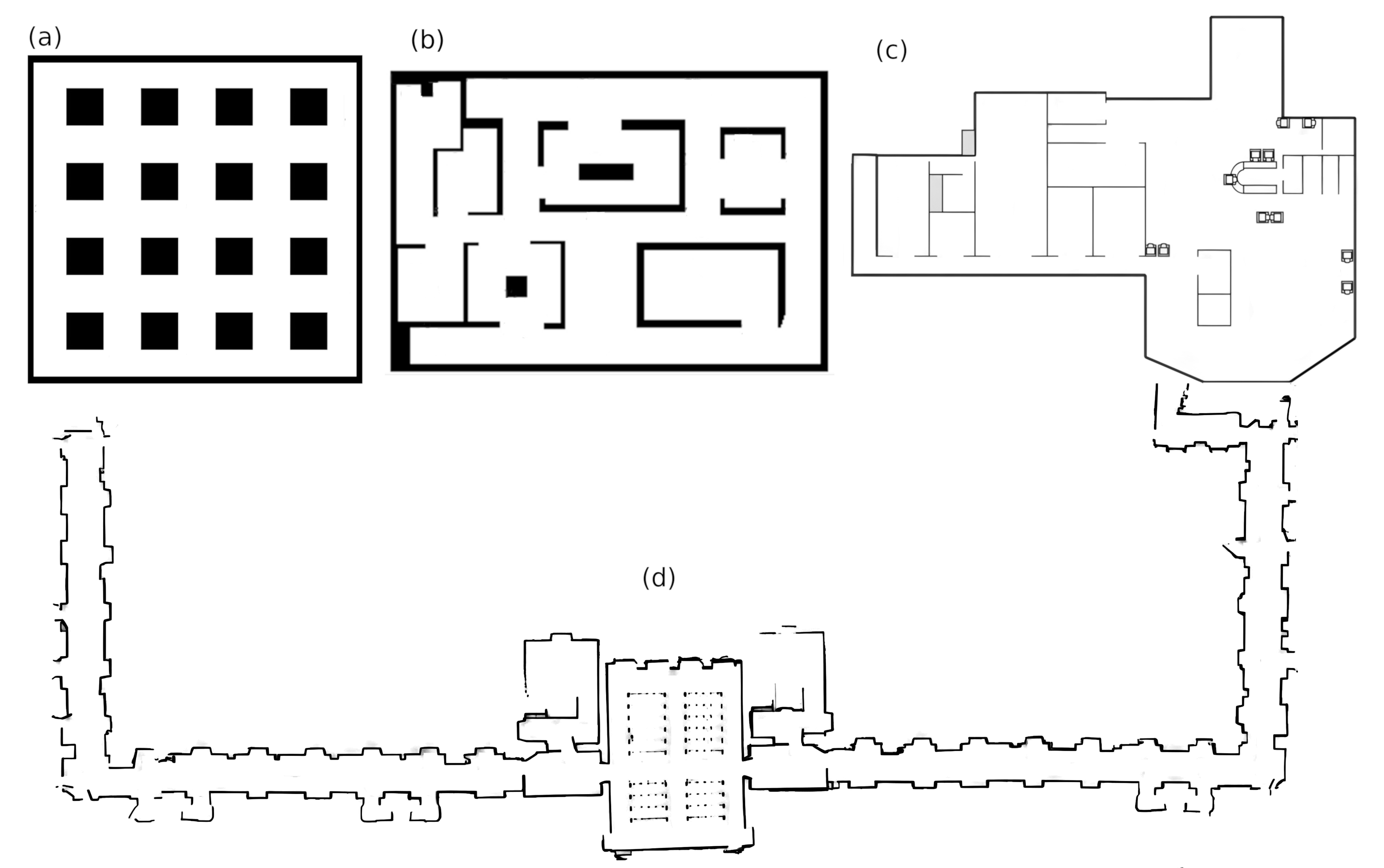}
    \centering
    \vspace{-3mm}
    \caption{The four maps used in testing: a) ``grid'' b) ``example'' c) ``cumberland'' d) ``DIAG\_floor1'' (not to scale)}
     \label{fig:maps}
     \vspace{-6mm}
\end{figure*}

In \textit{ROS patrolling sim}, SUNS and MNS plus leading multi-robot patrol strategies State Exchange Bayesian Strategy (SEBS)~\cite{Portugal2013}, Concurrent Bayesian Learning Strategy (CBLS)~\cite{Portugal2016}, and auction-based Dynamic Task Assignment (DTAP)~\cite{Farinelli2017} were simulated on four different maps  provided in the simulator (``grid'', ``example'', ``cumberland'', and ``DIAG\_floor1'', shown in Figure~\ref{fig:maps}).  These strategies were selected as baselines as they are fully decentralized and scenario agnostic, supported in \textit{ROS patrolling sim}, and promise leading levels of performance. Team sizes of 1, 2, 4, 6, 8, and 12 agents were tested. Each scenario, i.e. combination of strategy, map, and team size, was simulated five times with five different sets of starting positions (giving 120 total simulation runs per strategy) for six hours per run, as six hours is the duration recommended in \cite{Ward2023} to allow for adversarial performance estimates to stabilize. Each simulation run returned vertex idlenesses and agent positions logged in one-second increments. These logs were then used to estimate adversarial performance as described in \cite{Ward2023}, and to assess idleness minimization performance over long time periods. We measure mean vertex idleness as the mean of means of instantaneous idlenesses of all vertices of the patrol graph across the entire simulation time, and mean-maximum vertex idleness as the mean of the maximum instantaneous idlenesses recorded for each vertex.

\vspace{-1mm}

\subsubsection{Intelligent adversary model}

As previously mentioned, to assess performance against an intelligent adversary, we use the model proposed in~\cite{Ward2023}. The goal of this adversary is to spend a fixed attack duration at a target vertex without being visited by a patrol agent. This adversary model has knowledge of the patrol graph, and knows the physical locations of all patrol agents at all times (and therefore can infer the instantaneous idlenesses of the vertices of the patrol graph). It has access to no other information, such as the decision-making processes of the agents. While observing the patrol graph and agents, the adversary learns an ``expected likelihood of success'' function for each vertex, where the inputs are vertex idleness, the distances of the patrol agents to the vertex, and the velocities of patrol agents towards the vertex. It also learns a similar function of ``expected likelihood of state'' to predict the probability of a given state of the patrol system occuring in any given timestep. From these, it selects a discretized set of ``attack states'' to maximise its probability of successfully attacking each node within a given time window. This is then tested on unseen patrol data to measure performance.
\vspace{-0mm}

\subsubsection{Imperfect communication examination}

To examine the effects of imperfect communication on our proposed strategies, in \textit{MAGESim} we implemented a communication model whereby agents would reject incoming messages with some probability. This probability, which we varied from $0\%$ to $100\%$, added an extra dimension to the same parameter space as was used in our main performance analysis --- so, for this, we varied map, team size, and message failure rate for SUNS and MNS. As before, we simulated five runs for each scenario considered, albeit for only one hour instead of six as we were not considering adversarial performance. We selected SEBS as our baseline for comparison, as it is known to be fault-tolerant~\cite{Portugal2013}.
\vspace{-3mm}
\section{Results}
\vspace{-2mm}

\subsection{Idleness minimization}

The average mean and mean-maximum instantaneous vertex idlenesses for all scenarios are shown in Tables \ref{table:grid}--\ref{table:diag}. To compare performance across all scenarios, relative mean and mean-maximum idlenesses were calculated by dividing by the best performance recorded in each scenario across all strategies. These values, averaged across all simulation runs for scenarios, are shown in Table \ref{table:idleness_results}. To determine statistical significance of the apparent differences in idleness minimization performance, we performed a Kruskal-Wallis test followed by a multiple pairwise Dunn's test with resultant $p$-values adjusted using the Holm-Bonferroni method on the results from all simulation runs\footnote[8]{Kruskal-Wallis into Dunn's test is a standard process for multiple pairwise comparison of non-normal distributions, and applying Holm-Bonferroni corrections allows for a reduction in the probability of false-positive errors following multiple comparisons.}. The results of this testing are shown in Tables \ref{table:avg_idleness_pvalues} and \ref{table:max_idleness_pvalues}, where $p<0.05$ indicates that the differences in observed performance are statistically significant.

\begin{table*}[h!]
    \centering
    \caption{Average mean vertex idleness and mean-max vertex idleness for ``grid'' map \\ (best performances in bold)}
    \label{table:grid}
    \begin{tabular}{r||rr||rr||rr||rr||rr}
        ~ & \multicolumn{2}{c||}{DTAP} & \multicolumn{2}{c||}{SEBS} & \multicolumn{2}{c||}{CBLS} & \multicolumn{2}{c||}{SUNS} & \multicolumn{2}{c}{MNS}\\
        ~ & mean & max & mean & max & mean & max & mean & max & mean & max \\ \hline
        1 & 115.4 & 255.6 & 115.2 & 270.7 & 230.3 & 928.1 & \textbf{108.3} & \textbf{238.1} & 119.3 & 252.4 \\ 
        2 & \textbf{55.7} & \textbf{150.5} & 58.1 & 216.2 & 102.2 & 525.1 & 57.8 & 220.3 & 58.2 & 214.0 \\ 
        4 & 31.3 & 125.1 & 37.8 & 135.9 & 51.2 & 320.3 & \textbf{29.7} & \textbf{105.0} & 29.7 & 105.1 \\ 
        6 & 24.5 & 125.8 & \textbf{19.7} & \textbf{74.6} & 29.9 & 221.8 & 20.0 & 80.0 & 19.9 & 78.5 \\ 
        8 & 21.9 & 134.7 & \textbf{15.1} & \textbf{70.1} & 21.3 & 178.7 & 15.5 & 75.6 & 15.4 & 77.1 \\ 
        12 & 27.6 & \textbf{70.6} & 12.1 & 108.9 & 14.3 & 143.6 & \textbf{11.5} & 101.6 & 11.9 & 118.6 \\ 
    \end{tabular}
    
\end{table*}

\begin{table*}[h!]
    \centering
    \caption{Average mean vertex idleness and mean-max vertex idleness for ``example'' map \\ (best performances in bold)}
    \label{table:example}
    \begin{tabular}{r||rr||rr||rr||rr||rr}
        ~ & \multicolumn{2}{c||}{DTAP} & \multicolumn{2}{c||}{SEBS} & \multicolumn{2}{c||}{CBLS} & \multicolumn{2}{c||}{SUNS} & \multicolumn{2}{c}{MNS}\\
        ~ & mean & max & mean & max & mean & max & mean & max & mean & max \\ \hline
        1 & \textbf{187.9} & 428.8 & 224.7 & 629.2 & 464.9 & 1697.5 & 200.0 & 436.5 & 196.6 & \textbf{426.1} \\ 
        2 & 107.4 & \textbf{291.1} & 111.2 & 443.5 & 368.1 & 1593.4 & 101.2 & 359.4 & \textbf{99.3} & 350.8 \\ 
        4 & 55.9 & 208.5 & 52.4 & 221.7 & 219.2 & 1293.4 & 48.1 & 188.7 & \textbf{47.5} & \textbf{186.6} \\ 
        6 & 32.7 & 157.9 & \textbf{30.2} & 142.6 & 161.9 & 1007.9 & 30.9 & 134.5 & 30.3 & \textbf{124.6} \\ 
        8 & 32.7 & 229.2 & \textbf{20.8} & 103.9 & 79.5 & 733.6 & 22.6 & 102.6 & 22.0 & \textbf{99.3} \\ 
        12 & 26.3 & 144.9 & \textbf{14.3} & 100.9 & 42.4 & 420.5 & 15.7 & 131.8 & 14.7 & \textbf{100.8} \\ 
    \end{tabular}
    
\end{table*}

\begin{table*}[h!]
    \centering
    \caption{Average mean vertex idleness and mean-max vertex idleness for ``cumberland'' map \\ (best performances in bold)}
    \label{table:cumberland}
    \begin{tabular}{r||rr||rr||rr||rr||rr}
        ~ & \multicolumn{2}{c||}{DTAP} & \multicolumn{2}{c||}{SEBS} & \multicolumn{2}{c||}{CBLS} & \multicolumn{2}{c||}{SUNS} & \multicolumn{2}{c}{MNS}\\
        ~ & mean & max & mean & max & mean & max & mean & max & mean & max \\ \hline
        1 & 316.4 & 1305.1 & 292.3 & 793.5 & 877.2 & 3124.4 & 284.0 & 766.1 & \textbf{280.9} & \textbf{695.8} \\ 
        2 & 153.8 & \textbf{471.5} & 170.2 & 632.1 & 609.2 & 2808.5 & 152.8 & 606.4 & \textbf{149.7} & 537.1 \\ 
        4 & \textbf{69.1} & 323.8 & 88.4 & 388.4 & 396.8 & 1888.7 & 72.6 & 324.8 & 70.5 & \textbf{293.0} \\ 
        6 & 49.5 & 299.2 & 51.9 & 246.5 & 253.0 & 1758.3 & 46.8 & 239.6 & \textbf{45.4} & \textbf{205.5} \\ 
        8 & 36.5 & 245.0 & 41.3 & 258.4 & 221.4 & 1764.5 & 35.3 & 209.6 & \textbf{34.3} & \textbf{180.0} \\ 
        12 & 27.9 & \textbf{224.1} & 26.7 & 203.4 & 73.7 & 710.9 & 26.2 & 244.0 & \textbf{26.0} & 237.0 \\ 
    \end{tabular}
    
\end{table*}

\begin{table*}[h!]
    \centering
    \caption{Average mean vertex idleness and mean-max vertex idleness for ``DIAG\_floor1'' map \\ (best performances in bold)}
    \label{table:diag}
    \begin{tabular}{r||rr||rr||rr||rr||rr}
        ~ & \multicolumn{2}{c||}{DTAP} & \multicolumn{2}{c||}{SEBS} & \multicolumn{2}{c||}{CBLS} & \multicolumn{2}{c||}{SUNS} & \multicolumn{2}{c}{MNS}\\
        ~ & mean & max & mean & max & mean & max & mean & max & mean & max \\ \hline
        1 & \textbf{347.6} & \textbf{892.6} & 525.7 & 1729.9 & 1289.7 & 4493.0 & 359.4 & 892.7 & 370.2 & 955.5 \\ 
        2 & \textbf{181.9} & \textbf{633.3} & 238.6 & 958.5 & 769.9 & 3401.0 & 214.9 & 830.6 & 204.0 & 778.3 \\ 
        4 & \textbf{82.7} & \textbf{339.7} & 115.4 & 588.5 & 586.2 & 3145.6 & 105.6 & 451.7 & 96.7 & 442.0 \\ 
        6 & \textbf{57.3} & \textbf{264.1} & 73.8 & 464.0 & 377.6 & 2243.9 & 67.5 & 324.1 & 60.8 & 298.3 \\ 
        8 & 45.5 & 294.1 & 50.7 & 297.2 & 272.3 & 1887.0 & 49.2 & 276.8 & \textbf{44.3} & \textbf{233.8} \\ 
        12 & 55.3 & 660.1 & 32.2 & 234.0 & 178.0 & 1352.5 & 34.4 & 238.9 & \textbf{30.9} & \textbf{190.1} \\ 
    \end{tabular}
    
\end{table*}

\begin{table*}[h!]
    \centering
    \caption{Mean and standard deviation of mean and mean-maximum relative idleness results \\ (best performances in bold)}
    \label{table:idleness_results}
    \begin{tabular}{c||c c||c c||c c||c c||c c}
        ~ & \multicolumn{2}{c||}{DTAP} & \multicolumn{2}{c||}{SEBS} & \multicolumn{2}{c||}{CBLS} & \multicolumn{2}{c||}{SUNS} & \multicolumn{2}{c}{MNS}\\
         & $\mu$ & $\sigma$ & $\mu$ & $\sigma$ & $\mu$ & $\sigma$ & $\mu$ & $\sigma$ & $\mu$ & $\sigma$ \\ \hline
         Mean idleness & 1.26 & 0.35 & 1.19 & 0.16 & 4.10 & 1.94 & 1.10 & 0.10 & \textbf{1.07} & 0.06 \\
         Mean-max idleness  & 1.55 & 0.60 & 1.53 & 0.36 & 6.60 & 2.85 & 1.36 & 0.26 & \textbf{1.28} & 0.26
    \end{tabular}
\end{table*}

\newpage

    \begin{table}[H]
        \setlength\tabcolsep{2.5pt}
        \centering
        \caption{$p$-values for mean idleness minimization performance comparison (significant values in bold)}
        \vspace{-2mm}
        \label{table:avg_idleness_pvalues}
        \begin{tabular}{c|c c c c c}
                 & DTAP  & SEBS  & CBLS  & SUNS  & MNS   \\ \hline
            DTAP &   -   & 0.439 & \textbf{0.000} & \textbf{0.034} & \textbf{0.000} \\ 
            SEBS & 0.439 &   -   & \textbf{0.000} & \textbf{0.004} & \textbf{0.000} \\ 
            CBLS & \textbf{0.000} & \textbf{0.000} &   -   & \textbf{0.000} & \textbf{0.000} \\ 
            SUNS & \textbf{0.034} & \textbf{0.004} & \textbf{0.000} &   -   & 0.175 \\
            MNS  & \textbf{0.000} & \textbf{0.000} & \textbf{0.000} & 0.175 &   -  
        \end{tabular}
        \vspace{-4mm}
    \end{table}%

    \begin{table}[H]
        \setlength\tabcolsep{2.5pt}
        \centering
        \caption{$p$-values for mean-maximum idleness minimization performance comparison (significant values in bold)}
        \vspace{-2mm}
        \label{table:max_idleness_pvalues}
        \begin{tabular}{c|c c c c c}
                 & DTAP  & SEBS  & CBLS  & SUNS  & MNS   \\ \hline
            DTAP &   -   & \textbf{0.009} & \textbf{0.000} & 0.728 & 0.173 \\ 
            SEBS & \textbf{0.009} &   -   & \textbf{0.000} & \textbf{0.003} & \textbf{0.000} \\ 
            CBLS & \textbf{0.000} & \textbf{0.000} &   -   & \textbf{0.000} & \textbf{0.000} \\ 
            SUNS & 0.728 & \textbf{0.003} & \textbf{0.000} &   -   & 0.242 \\
            MNS  & 0.173 & \textbf{0.000} & \textbf{0.000} & 0.242 &   -  
        \end{tabular}
        \vspace{-4mm}
    \end{table} 
    
\newpage
\subsection{Adversarial performance}

\begin{figure}[H]
    \centering
    \normalsize
    \includegraphics[width=0.9\linewidth]{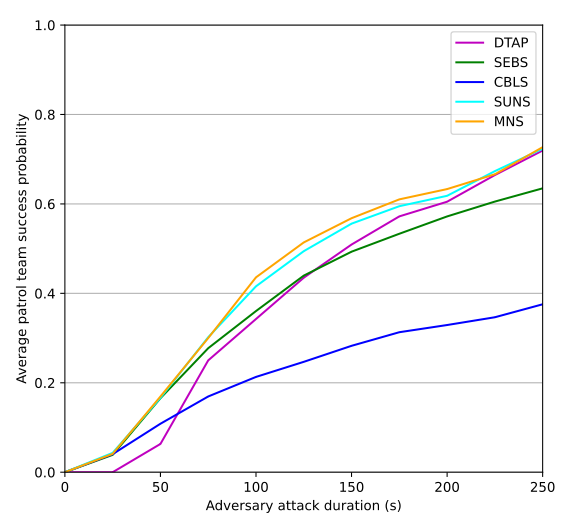}
    \centering
    \vspace{-5mm}
    \caption{Patrol team success probability against adversary as attack duration varies, averaged across all scenarios}
    \label{fig:adversarial_results}
    \vspace{-5mm}
\end{figure}

Tables \ref{table:adversarial_results} and \ref{table:adversarial_pvalues} and Figure \ref{fig:adversarial_results} show the results from our adversarial analysis, including statistical significance of difference in performance between strategies as determined by the method presented by \cite{Hristova2023}. In Table \ref{table:adversarial_results}, patrol team success probability $p(s)$ is the probability that the adversary is not able to successfully remain undetected at its target vertex for at least its attack duration after attacking, and difference-from-best is the difference in performance between a given strategy and the best performing strategy for a given scenario and adversary attack duration. 

\vspace{-3mm}

\vspace{-3mm}
\begin{table}[H]
    \setlength\tabcolsep{2.5pt}
    \centering
    \caption{Mean patrol team success probability in adversarial scenarios $p(s)$ and mean difference-from-best $\overline{\Delta}$}
    \vspace{-2mm}
    \label{table:adversarial_results}
    \begin{tabular}{c||c c c c c}
         ~ & DTAP & SEBS & CBLS & SUNS & MNS \\ \hline
         $p(s)$ & 0.378 & 0.375 & 0.220 & 0.417 & 0.424 \\
         $\overline{\Delta}$  & 0.086 & 0.090 & 0.244 & 0.047 & 0.040
    \end{tabular}
\end{table}%

\begin{table}[H]
    \setlength\tabcolsep{2.5pt}
    \centering
    \caption{$p$-values for adversarial performance comparison (significant values in bold)}
    \vspace{-2mm}
    \label{table:adversarial_pvalues}
    \begin{tabular}{c|c c c c c}
             & DTAP  & SEBS  & CBLS  & SUNS  & MNS   \\ \hline
        DTAP &   -   & \textbf{0.000} & \textbf{0.000} & \textbf{0.000} & \textbf{0.000} \\ 
        SEBS & \textbf{0.000} &   -   & \textbf{0.000} & \textbf{0.000} & \textbf{0.000} \\ 
        CBLS & \textbf{0.000} & \textbf{0.000} &   -   & \textbf{0.000} & \textbf{0.000} \\ 
        SUNS & \textbf{0.000} & \textbf{0.000} & \textbf{0.000} &   -   & 0.154 \\
        MNS  & \textbf{0.000} & \textbf{0.000} & \textbf{0.000} & 0.154 &   -  
    \end{tabular}
    \vspace{-2mm}
\end{table}

\subsection{Imperfect communication}

Table \ref{table:communication} shows the results from our examination of the effect of imperfect communication on performance. To calculate these, the mean idlenesses of each of the five runs for each failure probability $p(f)$ for each scenario were averaged, and then normalized to the average mean idleness for $p(f) = 0$ for that scenario. These values were then averaged across the four maps used to generate the tables shown, giving relative idleness against $p(f)$ for $2-12$ agents.

\begin{table}[H]
\caption{Average relative mean idleness against communication failure probability $p(f)$ averaged over all maps}
\label{table:communication}
\begin{minipage}{.48\columnwidth}
    \centering
    \caption*{2 agents}
    \vspace{-2mm}
    \setlength\tabcolsep{2.5pt}
    \begin{tabular}{r|lll}
        $p(f)$ & SEBS & SUNS & MNS \\ \hline
        0\% & 1.00 & 1.00 & 1.00 \\ 
        25\% & 1.05 & 1.04 & 1.03 \\ 
        50\% & 1.09 & 1.05 & 1.06 \\ 
        75\% & 1.15 & 1.12 & 1.10 \\ 
        80\% & 1.16 & 1.12 & 1.13 \\ 
        90\% & 1.20 & 1.17 & 1.23 \\ 
        100\% & 1.25 & 1.28 & 1.29 \\ 
    \end{tabular}
\end{minipage}
\hfill
\begin{minipage}{.48\columnwidth}
    \centering
    \caption*{4 agents}
    \vspace{-2mm}
    \setlength\tabcolsep{2.5pt}
    \begin{tabular}{r|lll}
        $p(f)$ & SEBS & SUNS & MNS \\ \hline
        0\% & 1.00 & 1.00 & 1.00 \\ 
        25\% & 1.05 & 1.05 & 1.06 \\ 
        50\% & 1.14 & 1.13 & 1.13 \\ 
        75\% & 1.28 & 1.26 & 1.28 \\ 
        80\% & 1.33 & 1.30 & 1.35 \\ 
        90\% & 1.47 & 1.50 & 1.49 \\ 
        100\% & 1.68 & 1.71 & 1.70 \\ 
    \end{tabular}
\end{minipage}

\vspace{2mm}

\begin{minipage}{.48\columnwidth}
    \centering
    \caption*{6 agents}
    \vspace{-2mm}
    \setlength\tabcolsep{2.5pt}
    \begin{tabular}{r|lll}
        $p(f)$ & SEBS & SUNS & MNS \\ \hline
        0\% & 1.00 & 1.00 & 1.00 \\ 
        25\% & 1.08 & 1.07 & 1.08 \\ 
        50\% & 1.20 & 1.19 & 1.20 \\ 
        75\% & 1.44 & 1.40 & 1.42 \\ 
        80\% & 1.51 & 1.51 & 1.53 \\ 
        90\% & 1.69 & 1.76 & 1.75 \\ 
        100\% & 1.78 & 1.95 & 1.97 \\ 
    \end{tabular}
\end{minipage}
\hfill
\begin{minipage}{.48\columnwidth}
    \centering
    \caption*{8 agents}
    \vspace{-2mm}
    \setlength\tabcolsep{2.5pt}
    \begin{tabular}{r|lll}
        $p(f)$ & SEBS & SUNS & MNS \\ \hline
        0\% & 1.00 & 1.00 & 1.00 \\ 
        25\% & 1.09 & 1.09 & 1.11 \\ 
        50\% & 1.26 & 1.22 & 1.26 \\ 
        75\% & 1.56 & 1.49 & 1.54 \\ 
        80\% & 1.65 & 1.60 & 1.65 \\ 
        90\% & 1.96 & 1.99 & 2.04 \\ 
        100\% & 2.29 & 2.29 & 2.23 \\ 
    \end{tabular}
\end{minipage}

\vspace{2mm}

\begin{minipage}{.48\columnwidth}
    \centering
    \caption*{12 agents}
    \vspace{-2mm}
    \setlength\tabcolsep{2.5pt}
    \begin{tabular}{r|lll}
        $p(f)$ & SEBS & SUNS & MNS \\ \hline
        0\% & 1.00 & 1.00 & 1.00 \\ 
        25\% & 1.13 & 1.15 & 1.17 \\ 
        50\% & 1.34 & 1.31 & 1.36 \\ 
        75\% & 1.79 & 1.67 & 1.75 \\ 
        80\% & 1.95 & 1.82 & 1.91 \\ 
        90\% & 2.51 & 2.40 & 2.46 \\ 
        100\% & 2.80 & 2.87 & 2.65 \\ 
    \end{tabular}
\end{minipage}
\hfill
\begin{minipage}{.48\columnwidth}
    \centering
    \caption*{Averaged over all teamsizes}
    \vspace{-2mm}
    \setlength\tabcolsep{2.5pt}
    \begin{tabular}{r|lll}
        $p(f)$ & SEBS & SUNS & MNS \\ \hline
        0\% & 1.00 & 1.00 & 1.00 \\ 
        25\% & 1.08 & 1.08 & 1.09 \\ 
        50\% & 1.21 & 1.18 & 1.20 \\ 
        75\% & 1.44 & 1.39 & 1.42 \\ 
        80\% & 1.52 & 1.47 & 1.52 \\ 
        90\% & 1.77 & 1.76 & 1.79 \\ 
        100\% & 1.96 & 2.02 & 1.97 \\ 
    \end{tabular}
\end{minipage}
\end{table}


\section{Discussion}
\label{section:discussion}

Our results shows that both SUNS and MNS offer significantly better ($p<0.05$) performance in mean idleness minimization than any of the literature strategies tested, while not performing significantly differently ($p>0.05$) from each other (Table~\ref{table:avg_idleness_pvalues}). Mean-maximum idleness minimization showed less improvement, with both SUNS and MNS out-performing SEBS but showing no significant improvement over DTAP (Table~\ref{table:max_idleness_pvalues}). This is likely due to the wider variances in performance of mean-max versus mean idleness making it harder to confidently separate the distributions of observed performance. Adversarial performance similarly showed both SUNS and MNS outperforming other strategies tested, while showing no significant difference between each other. Our analysis of adversarial performance is limited partly due to the limitations of the adversary models used, as discussed later, but also because the curves of patrol team success probability against adversary attack duration vary widely across different scenarios. This makes it challenging to draw strong conclusions about performance. However, we suggest that the apparent improved performance of SUNS and MNS in adversarial settings can be attributed to the improved idleness performances --- as both SUNS and MNS are fully deterministic, they cannot be assumed to have any means to ``suprise'' a sufficiently intelligent attacker, but more efficient monitoring of a patrol graph will leave fewer possible windows in which an adversary could attack, thus improving adversary detection performance.

Our examination of imperfect communication found that SUNS and MNS showed very similar performance degradation to SEBS as communication became increasingly sporadic. SEBS has previously been noted to exhibit graceful degradation of performance as probability of communication failure $p(f)$ increases~\cite{Portugal2013}, which is consistent with our observations, and both SUNS and MNS also displayed this behavior. Performance for all three strategies considered was also found to degrade faster with $p(f)$ for larger teamsizes, which can be explained by the increased reliance on communication to maintain efficient behavior and avoid inter-robot interference in larger teams. Performance was observed to degrade fairly gradually up to $p(f) = 50\%$ and then to degrade significantly from there, suggesting robustness to moderate communication failure rates.

We attribute the poor performance of CBLS in our tests to how we measure mean idleness on a patrol graph. Previous tests~\cite{Portugal2018} of CBLS maintain a list of idlenesses of vertices as they are visted, and then average over that list. However, this approach under-weights cases where a vertex goes a long period without a visit (which we expose in our measures of mean-max idleness), meaning that a strategy that prioritizes frequent visiting of a subset of the patrol graph will give results that do not accurately reflect the vertices that may have been left unvisited for long periods. Our measure of mean idleness does not have this issue, as the idleness of every vertex at every time step is logged. 

The improved performance of MNS compared to existing strategies, while achieving a comparable (not significantly different) level of performance to the more complex SUNS, suggests that neither complex utility functions (in the case of SUNS) or complex inter-agent coordination mechanisms (in the case of DTAP) are necessary to achieve a high level of performance in distributed multi-robot patrol. However, the most notable difference between MNS and SEBS is the utility function -- SEBS simply uses vertex idleness divided by the distance to said vertex, while MNS uses three trained neurons. This suggests that some level of sophistication of utility function is highly beneficial, but complex neural networks are not necessary to achieve high-level performance. The fact that MNS performs comparably well to SUNS, which considers a vertex's neighbors when determining its utility, also suggests that this vertex neighbor information is not necessary for high-performing patrol strategies --- this is reinforced by the tendency of high-performing literature strategies to also discard vertex neighbor information when calculating vertex utilities (GBS~\cite{Portugal2013_2}, SEBS~\cite{Portugal2013_2}, CBLS~\cite{Portugal2016}, DTAG~\cite{Farinelli2017}, DTAP~\cite{Farinelli2017}, and ER~\cite{Yan2016} all consider each vertex in isolation), suggesting that previous authors have come to the same conclusion while developing their patrol strategies.

Despite the ability of GNNs to act as learnable coordination policies in multi-robot teams, here we have instead used an extremely simple coordination mechanism alongside our learned single-agent controllers. During initial testing, we found that our coordination mechanism generally outperformed simple GNN-based policies, suggesting that, for this problem, simple coordination mechanisms to allow agents to effectively ``soft-partition'' the environment can be extremely effective (i.e. when each robot tends to stay in a set of vertices which other agents will not often visit, and these regions change slowly over time) . This ``soft-partitioning'' behavior can be observed in many literature patrol strategies, and as ``harder'' partitioning can give optimal idleness minimization behavior in many cases~\cite{Chevaleyre2004} it is unsurprising that successful decentralized strategies tend to mimic this behavior.

\label{section:limitations}
The most significant limitation of this work is that it was carried out entirely in simulation. While \textit{ROS patrolling sim} is a well-established simulator, it still presents an idealized environment. As such, we cannot be certain that conclusions drawn in this work could apply directly to real-world deployments. Our plans for future research include validating both SUNS and MNS on real robots, but due to the impracticality of performing real-world tests on as large a scale as was presented in this work, strong conclusions about relative strategy performance may be difficult to draw in a research context. 

A second limitation is the methods used to measure performance in both idleness minimization and adversarial settings. The ``intelligent adversary'' used is not necessarily representative of a real hostile actor, as more sophisticated models may be able to achieve better performance, or leverage additional information. All patrol strategies considered in this work are deterministic, meaning that in principle a sufficiently sophisticated attacker model could achieve perfect performance against them. Additionally, our measure of mean idleness may also be less directly relevant to specific real deployments, as it weights all vertices of the patrol graph equally.

\section{Conclusions}
In this work, we present two lightweight neural network-based strategies for decentralized multi-robot patrol. Both strategies are found to outperform leading literature strategies in idleness minimization and in defense against an intelligent adversary model. Both strategies are additionally found to offer a high level of resiliency to imperfect inter-agent communication, with only slight performance degradation for moderate communication failure rates. By considering the differences in performance and architecture between our new strategies and existing strategies, we suggest that complex utility functions or inter-robot coordination mechanisms are not necessary for high levels of performance in this problem.

\section*{Acknowledgment}
This work was supported by the Engineering and Physical Sciences Research Council (EPSRC) under grant no. EP/S021795/1 and by the Royal Academy of Engineering under the Research Fellowship program.

\bibliographystyle{IEEEtran}
\bibliography{sample-bibliography}

\end{document}